%% file: elsarticle-template-num.tex
\documentclass[preprint,review,12pt]{elsarticle}

\usepackage{amssymb}
\usepackage[dvipsnames]{xcolor}

\input{math_commands.tex}

\newcommand{\mpxgat}{\emph{MPXGAT}}

\newcommand{\mpxgatHmodel}{\emph{MPXGAT-H}}

\newcommand{\mpxgatVmodel}{\emph{MPXGAT-V}}

\newcommand{\multiplexsagebf}{\textbf{\emph{MultiplexSAGE}}}
\newcommand{\multiplexsage}{\emph{MultiplexSAGE}}
    
\newcommand{\gatnebf}{\textbf{\emph{GATNE}}}
\newcommand{\gatne}{\emph{GATNE}}
    
\newcommand{\graphsagebf}{\textbf{\emph{GraphSAGE}}}
\newcommand{\graphsage}{\emph{GraphSAGE}}

\newcommand{\gat}{\emph{GAT}}

\newcommand{\gatv}{\emph{GAT-V}}

\journal{}

\begin{document}

\begin{frontmatter}

\title{MPXGAT: An Attention based Deep Learning Model for Multiplex Graphs Embedding}

\author[1]{Marco Bongiovanni}
\ead{marco.bongiovanni94@gmail.com}
\author[2]{Luca Gallo}
\author[3]{Roberto Grasso}
\author[4]{Alfredo Pulvirenti\corref{cor1}}
\ead{alfredo.pulvirenti@unict.it}
\affiliation[1]{organization={Department of Mathematics and Computer Science, University of Catania},
            addressline={Via Santa Sofia, 64}, 
            city={Catania},
            postcode={95125}, 
            country={Italy}}
\affiliation[2]{organization={Department of Network and Data Science, Central European University},
            addressline={Quellenstra{\ss}e 51}, 
            city={Vienna},
            postcode={1100}, 
            country={Austria}}
\affiliation[3]{organization={Department of Physics and Astronomy, University of Catania},
            addressline={Via Santa Sofia, 64}, 
            city={Catania},
            postcode={95125}, 
            country={Italy}}
\affiliation[4]{organization={Department of Clinical and Experimental Medicine, University of Catania},
            addressline={Via Santa Sofia, 64}, 
            city={Catania},
            postcode={95125}, 
            country={Italy}}

\cortext[cor1]{Corresponding author}

\begin{abstract}
Graph representation learning has rapidly emerged as a pivotal field of study. Despite its growing popularity, the majority of research has been confined to embedding single-layer graphs, which fall short in representing complex systems with multifaceted relationships. To bridge this gap, we introduce MPXGAT, an innovative attention-based deep learning model tailored to multiplex graph embedding. Leveraging the robustness of Graph Attention Networks (GATs), MPXGAT captures the structure of multiplex networks by harnessing both intra-layer and inter-layer connections. This dual exploitation facilitates accurate link prediction within and across the network's multiple layers. Our comprehensive experimental evaluation, conducted on various benchmark datasets, confirms that MPXGAT consistently outperforms state-of-the-art competing algorithms.

\end{abstract}

\begin{highlights}
\item We introduce MPXGAT, an attention-based model for embedding multiplex networks
\item MPXGAT is effective in predicting links across layers of real-world multiplexes
\item MPXGAT shows significant improvement over state-of-the-art competitors
\end{highlights}


\end{frontmatter}

\section{Introduction and Related Works}
    In the last decades, graphs have proved to be a fundamental mathematical tool to model various real-world complex systems.
    From transportation systems to power grids, from the network of our social relationships to that of neurons in our brains, complex networks are all around us. 
    Due to such ubiquity, network and graph theory have imposed themselves in many research fields, from engineering to physics, social science, and biology~\citep{newman2003structure,boccaletti2006complex,latora2017complex,newman2018networks}. 
    
    A topic that has recently received considerable interest in computer science is that of how to efficiently represent large-scale graphs~\citep{bacciu2020gentle,chen2020graph,grohe2020word2vec}. 
    Particularly, graph embedding methods, which consist in projecting the elements of a graph, i.e., vertices, edges, and motifs, to a low-dimensional vector space by preserving some of the graph properties, have shown to be very successful in graph representation~\citep{khoshraftar2022survey}.
    These embedding techniques are suitable for multiple applications, as they can be used in downstream learning tasks, including node classification~\citep{rong2019dropedge}, link prediction~\citep{lu2011link}, and community detection~\citep{fortunato2010community}.
    
    We can broadly categorize graph embedding methods into traditional graph embedding and graph neural networks (GNNs) based embedding methods~\citep{khoshraftar2022survey}.
    The first group consists of algorithms that represent graphs relying on techniques such as random walks~\citep{perozzi2014deepwalk,tang2015line,grover2016node2vec} and matrix factorization methods. 
    These shallow approaches have several drawbacks that limit their efficiency and effectiveness. First, they do not share any parameters between the nodes in the encoder function, which maps each node to a vector representation, making them statistically and computationally inefficient. Second, they ignore the node attributes during the encoding process, leading to a lower quality of the embeddings. Third, they are transductive, meaning they can only generate embeddings for the nodes seen during the training phase~\citep{hamilton2017representation}.
    
    Embedding methods based on GNNs go beyond such limitations using more sophisticated encoders accounting for the graph structure and the node attributes. 
    The key feature of a GNN is that the embedding of a node in the graph is obtained by \textit{aggregating} the embeddings of the node’s neighbors~\citep{khoshraftar2022survey}.
    In the last few years, a large variety of methods have been developed~\citep{chen2020graph,velivckovic2017graph,liu2020towards,gao2019graph}, with the most notable examples including Graph Convolutional Networks (GCNs)~\citep{kipf2016semi}, GraphSAGE \citep{hamilton2017inductive} and Graph Attention networks (GATs)~\citep{velivckovic2017graph}.

Although graphs are widely recognized for their versatility in modeling complex systems, traditional graph structures are limited to depicting a singular relationship type among system entities. This simplistic representation is insufficient for systems where entities engage in diverse interactions~\citep{latora2017complex}. For example, social systems often involve individuals connected through various relationships such as friendship, kinship, or professional collaboration, and they may communicate through multiple channels like direct contact, phone, or online platforms~\citep{szell2010multirelational, battiston2014structural}. Similarly, in biological systems, proteins may interact genetically, physically, or through spatial proximity~\citep{de2015structural}.
       
    To encapsulate the complexity of such systems, advanced mathematical structures like multidimensional and multiplex graphs are employed~\citep{boccaletti2014structure,berlingerio2011foundations,cozzo2018multiplex}. Multidimensional graphs, or edge-labeled multigraphs, feature vertices connected by edges of different labels, each signifying a distinct interaction type within the system. Multiplex graphs, on the other hand, consist of multiple interconnected layers, with each layer dedicated to a specific relationship type.
    
    The key distinction between multigraphs and multiplex networks lies in the representation of system units. In multigraphs, a unit is depicted as a single node linked to others through various connection types. Multiplex networks~\citep{de2013mathematical, battiston2014structural}, however, represent units as a set of nodes distributed across different layers, connected by inter-layer links. Typically, a unit is represented in only a subset of layers, indicating connections through certain relationship types. Furthermore, it is often the case that only a portion of the inter-layer links connecting the multiple representations of the same unit is known~\citep{cozzo2018multiplex}.
    
    Multi-relational network embedding is a challenging problem recently garnering significant research interest. 
    Various methods to embed networks with multiple types of nodes and links, such as multidimensional networks and multiplex networks, have been proposed. 
    Some of these methods are based on shallow embedding approaches~\citep{liu2017principled,zhang2018scalable,shi2018mvn2vec,gong2020heuristic}, while others use graph convolutional networks (GCNs)~\citep{yang2020multisage,ioannidis2020tensor,huang2020mr} or graph attention networks (GATs)~\citep{cen2019representation}. 
    However, none of these methods can solve the problem of predicting links between different layers of a multilayer network. 
    This problem is important when the network structure is incomplete or heterogeneous. 
    For example, some methods assume that all nodes are present in every layer, which is not realistic in many cases. 
    Other methods (such as \citep{behrouz2022anomaly}; \citep{10.1145/3511808.3557572}; \citep{qu2017attentionbased}) do not distinguish between intra-layer and inter-layer links, ignoring the diversity and complexity of multilayer networks. 
    Very recently, MultiplexSAGE, a generalization of GraphSAGE aimed at embedding multiplex networks by relaxing these hypothesis, have been proposed~\citep{gallo2023multiplexsage}, yet there is still a need for new methods that can address the problem of predicting inter-layer links in a general, flexible and more reliable way.

    The inter-layer link prediction problem finds many crucial applications: i) linking user identities across different online social networks (OSNs), an emerging task in social media that has attracted increasing attention in recent years~\citep{shu2017user}. User identity linkage finds potential impact in different domains, from recommendation systems to cybersecurity~\citep{tang2020interlayer}; 
    ii) identifying the same genes or proteins across different biological networks, such as gene expression, protein interaction, metabolic pathways, etc. This can help to discover the molecular mechanisms of diseases and to find potential drug targets~\citep{jain2023composed}; 
    iii) matching the same entities across different knowledge graphs~\citep{azmy2019matching}, which can help to enrich the semantic information and to improve query answering and reasoning capabilities. 
    The range of possible applications of inter-layer link prediction ultimately motivates the development of embedding algorithms for multiplex networks that are able to distinguish between intra-layer and inter-layer links, and to reconstruct both connectivity patterns. 
    
    Predicting inter-layer connections in multiplex networks differs from graph matching or alignment. Graph matching~\citep{mathon1979note} identifies isomorphisms between graphs, ensuring identical topology. In contrast, inter-layer prediction uncovers missing links between nodes representing the same unit across layers, with diverse connectivity. Global graph alignment~\citep{ma2020review} finds optimal correspondences between multiple graphs, considering both structure and node labels. However, inter-layer link prediction focuses on a specific bijection between two graphs, i.e., layers of a multiplex network. Also, while graph alignment is general, inter-layer prediction is specific to multiplex networks.
    
    In this paper, we introduce MPXGAT, an attention based deep learning model for multiplex graphs embedding.
    Our methodology, based on GATs, consists in embedding the nodes of a multiplex network by leveraging the information about their intra-layer and inter-layer connections, allowing for link prediction tasks both within the same layer and across different layers.
    We carry out a thorough experimental analysis on three benchmark datasets, showing that MPXGAT out-performs state-of-the-art competing algorithms.
    We conclude with an in-depth study of the model main features, proving how their use positively impacts the performance of the algorithm itself. 
    
    \section{Methods}
    \label{sec:methods}
    In this section, we provide details of our model, \mpxgat{}, which can embed multiplex networks, namely networks where multiple types of link exist.
    We first provide some preliminary notions, next, we introduce the mathematical formulation of the model, and finally, we describe the algorithmic implementation.
    
    \subsection{Preliminary Notions}

    We start by defining the basic concepts of simple and multiplex graphs
    A graph (network) is a mathematical structure consisting of a set of vertices (nodes) and a set of edges (links), where each edge represents a relation between a pair of vertices. 
    Simple graphs are able to capture only one kind of relationship, so they are limited when it comes to characterizing systems where multiple types of interactions coexist. 
    The need of a higher degree of expressiveness motivates the use of multiplex graphs. 
    We can define multiplex graphs as a graphs composed by two different sub-networks, which we refer to as horizontal and vertical.
    The horizontal network consists of a collection of simple graphs, called (horizontal) layers. 
    In this setup, each unit of a system can be represented as a node in one or more layers, with each layer referring to a certain type of relationship.
    We refer to the connections within a given layer as intra-layer links.
    The second sub-network, namely the vertical network, consists of a single-layer graph formed by the set of edges connecting nodes across different layers. 
    We assume that a node $i$ on a layer $\alpha$ can be connected to at most one node $j$ on another layer $\beta$, i.e., the two nodes represent the same unit of the system.
    Also, we assume that if node $i$ on layer $\alpha$ is connected to node $j$ on layer $\beta$, and if node $j$ is connected to node $k$ on layer $\gamma$, then nodes $i$ and $k$ are also connected. 
    Under these hypothesis, the vertical network consists in a collection of connected components, which can be either cliques or isolated nodes. 
    The edges in the vertical network will be referred to as inter-layer links.
    
\begin{figure}[h]
        \begin{center}
        \includegraphics[width=0.6\textwidth]{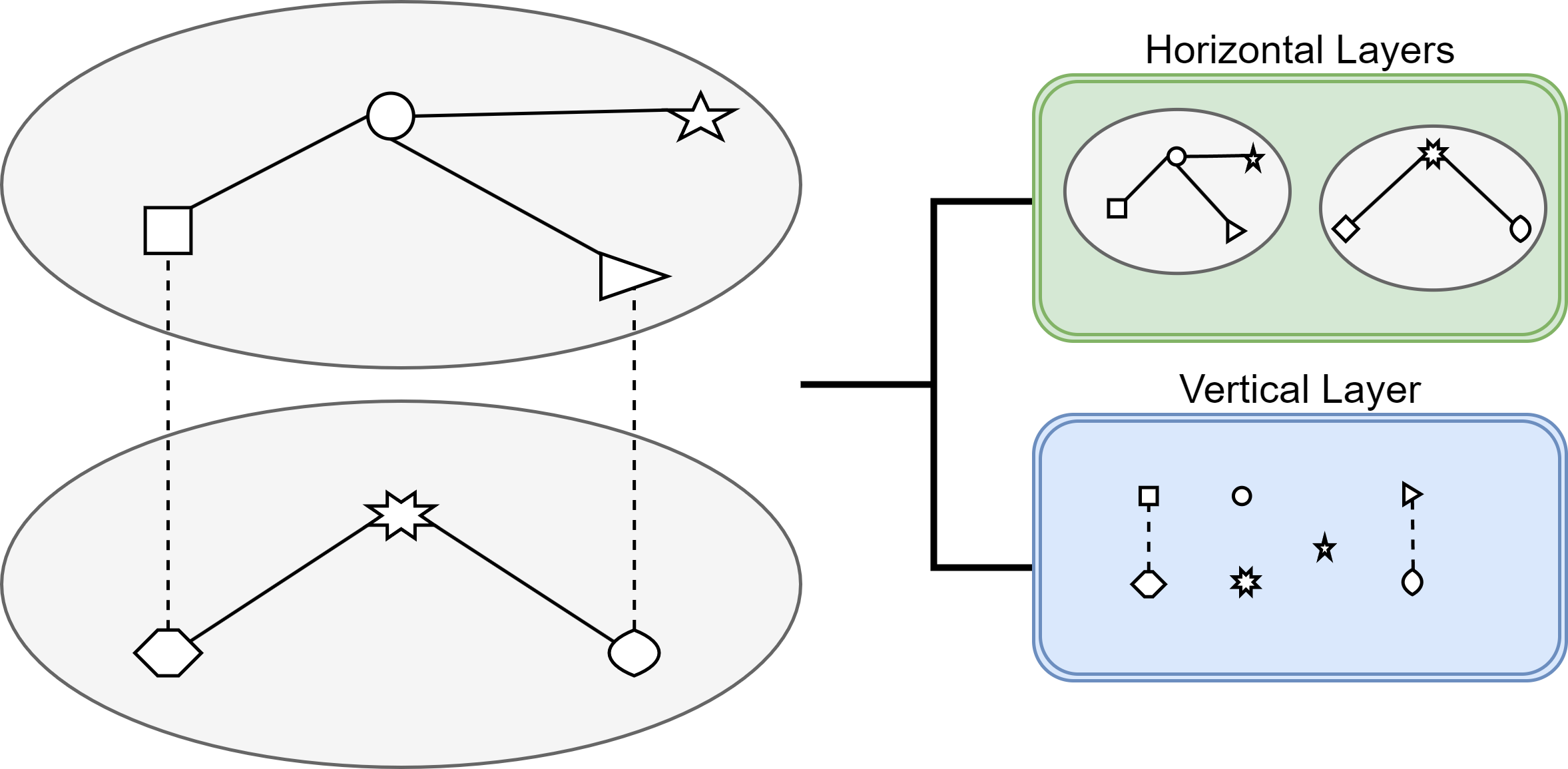}
        \end{center}
        \caption{A toy example of a multiplex network with 2 horizontal layers. The solid edges represent the intra-layer connections while the dashed edges are the inter-layer edges.}
    \label{fig:multiplex_graph_toy_example}
    \end{figure}

    Formally, a multiplex graph is a set $\sV$ of $N$ nodes that are connected through $L$ different layers. 
    We assume that $N_\alpha$ nodes are present in each layer $\alpha \in \{0, 1, \dots, L \}$, such that $N_1+ \dots + N_L = N$. 
    Each layer $\alpha$ is a graph $\gG_\alpha = \left(\sV_\alpha , \mathcal{E}_\alpha\right)$ where $\sV_\alpha \in \sV$ is the set of the $N_\alpha$ nodes and $\mathcal{E}_\alpha$ is the set of edges connecting them.
    The node sets for each layer are disjoint, i.e., $\sV_\alpha \cap \sV_\beta = \emptyset$ for $\alpha \neq \beta$, and their union makes the set of all nodes $\sV$.
    The set of intra-layer links for the horizontal network can be defined as $\mathcal{E}_\mathrm{intra} = \bigcup_{\alpha=1}^L\mathcal{E}_\alpha$.
    Hence, we define the horizontal network as $\gG_H = \left(\sV, \mathcal{E}_\mathrm{intra}\right)$.
    The vertical network can be instead defined as $\gG_V = \left(\sV, \mathcal{E}_\mathrm{inter}\right)$, where $\mathcal{E}_\mathrm{inter} = \left\{\left(i,j\right) \in \sV \times \sV \mid \exists \alpha, \beta \in \{0, 1, \dots, L \}\, s.t.\, \left(i,j\right) \in \mathcal{E}_\alpha \times \mathcal{E}_\beta \right\}$ is the set of inter-layer edges.
    Figure~\ref{fig:multiplex_graph_toy_example} shows an illustration of a multiplex graph.

    \begin{table}[t]
        \caption{Description of the variables involved in this work. Note that superscripts $\cdot^H$ and $\cdot^V$ are omitted for the sake of readability
        }
        \label{tab:math_eq_variables}
        \begin{center}
        \scalebox{0.73}{
            \begin{tabular}{ll}
                \multicolumn{1}{c}{\bf VARIABLE}  &\multicolumn{1}{c}{\bf DESCRIPTION}
                \\ \hline \hline \\
                $N \in \mathbb{N}$ & Number of nodes in the multiplex graph\\

                $\sV$ & Set of nodes in the multiplex graph\\

                $L \in \mathbb{N}$ & Total number of horizontal layers of the multiplex graph\\

                $\alpha \in \mathbb{N}$ & Index of the horizontal layer in the multiplex graph \\

                $N_\alpha$ & Number of nodes within the horizontal layer of index $\alpha$\\

                $\sV_\alpha$ & Set of nodes within the horizontal layer of index $\alpha$\\

                $\mathcal{E}_\alpha$ & Set of edges within the horizontal layer $\alpha$\\

                $\gG_\alpha$ & Layer of index $\alpha$ in the horizontal network\\

                $\mathcal{E}_\mathrm{intra}$ & Set of intra-layer edges for the horizontal network\\

                $\mathcal{E}_\mathrm{inter}$ & Set of inter-layer edges for the horizontal network\\

                $\gG_H$ & Horizontal network of the multiplex graph\\

                $\gG_V$ & Vertical network of the multiplex graph\\

                $F \in \mathbb{N}$ & Initial node feature dimensionality\\
    
                $F' \in \mathbb{N}$ & Final node feature dimensionality\\
    
                $i \in \mathbb{N}$ & Source node \\
    
                $j \in \mathbb{N}$ & Destination node \\
                
                $k \in \mathbb{N}$ & Source node horizontal layer's index \\
    
                $q \in \mathbb{N}$ & Destination node horizontal layer's index \\
    
                $\mathcal{N}_i$ & Set of nodes connected to node $i$ \\
                
                $\vh_i \in \R^F$ & Embedding of node $i$ \\
                
                $\vv \in \R^{F'} $ & Attention weight vector \\
                
                $\mW \in \R^{F' \times F}$ & Weight matrix, used to linearly transform the node embeddings\\
    
                $e_{i,j} \in \R$ & Attention coefficient between node $i$ and $j$ \\
    
                $\alpha_{i,j} \in \R$ & Normalized attention coefficient between nodes $i$ and $j$ \\
                
                $\vh_i\in \R^{F'}$ & Updated embedding of node $i$ after a forward pass \\
    
                $\sigma$ & Activation function (in our case LeakyReLU) \\
    
                $f$ & Function used to transform the horizontal embeddings \\ &to the same dimensional space of the vertical embedding\\
    
                $g$ & Function used to combine the data obtained from the vertical network\\& with the result of the application of the function $f$\\
                \hline
            \end{tabular}
            }
        \end{center}
    \end{table}
    
    \subsection{The MPXGAT General Framework}
    \label{subsec:math_formulation}
    The main idea of MPXGAT is to generate two separate embeddings for each node in two different phases. 
    In the first phase, each node is embedded according to the horizontal layers, where it has multiple types of relation with other nodes. 
    In the second phase, nodes are embedded according to the vertical network, where they are linked to their counterparts on different layers.
    In the following, we will use the superscript $\cdot^H$ when we refer to the horizontal network, while the superscript $\cdot^V$ will refer to the vertical one.
    The notation used, together with the most relevant variables of our model, are reported in Table~\ref{tab:math_eq_variables}. 
    
    \begin{figure}[t]
        \begin{center}
        \includegraphics[width=1\textwidth]{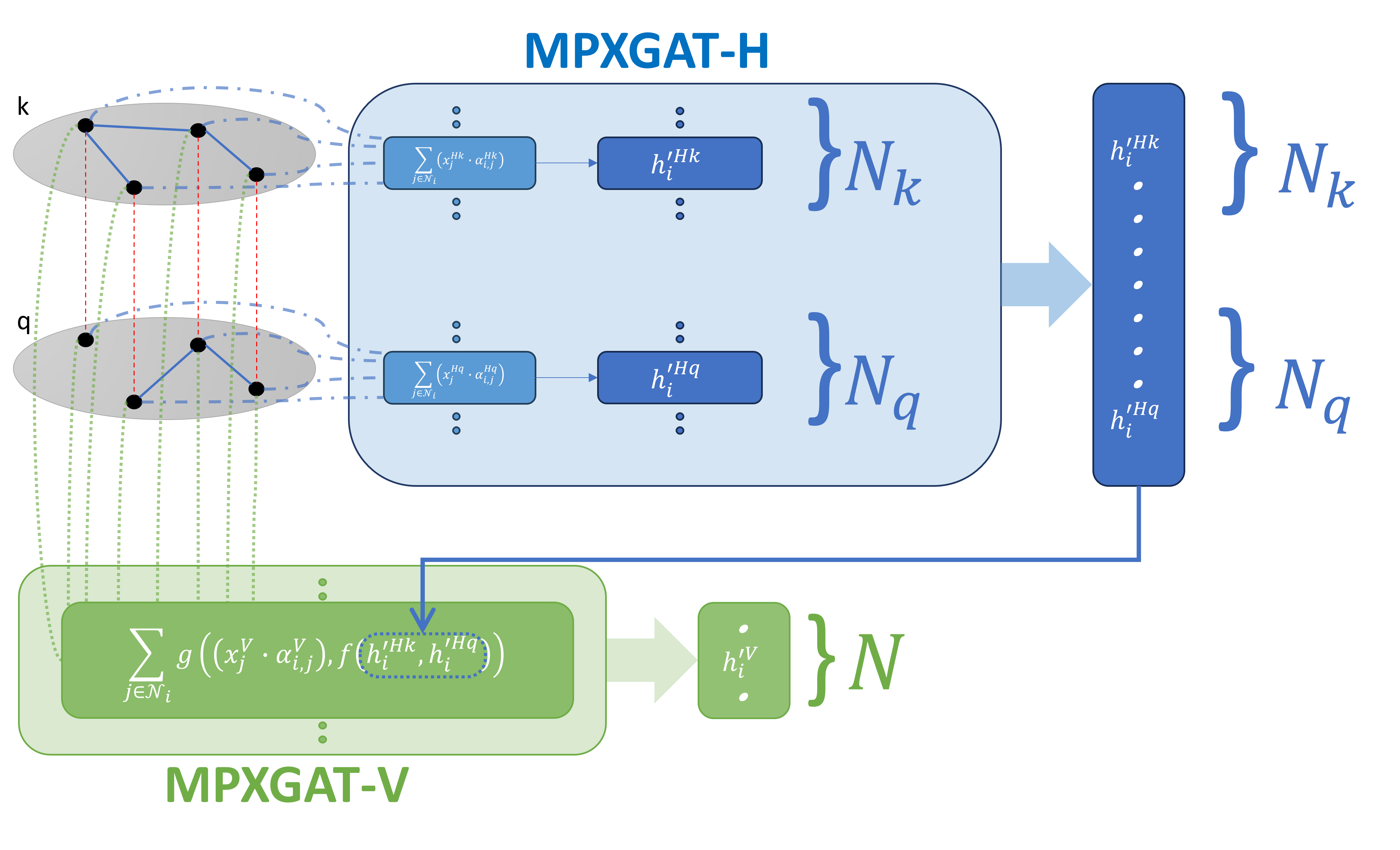}
        \end{center}
        \caption{The structure of the MPXGAT model. In this toy example, it is applied on a multiplex network with 2 horizontal layers where the solid blue edges represent the intra-layer connections while the dashed red edges are the inter-layer edges. The data is provided to the \mpxgatHmodel{} throughout the dot-and-dash blue lines. Once processed these are used to feed the \mpxgatVmodel{} together with the inter-layer links (the dotted green lines). The output of the model consists of both horizontal and vertical nodes embedding.}
    \label{fig:mpxgat_model_structure}
    \end{figure}
    
    Our algorithm uses two sub-models, one for each embedding phase: MPXGAT-H and MPXGAT-V (an illustration of the model structure is shown in Figure~\ref{fig:mpxgat_model_structure}). 
    In the most general framework, MPXGAT-H applies a series of GAT convolutional layers independently to each layer of the horizontal network
    Mathematically, the embedding is described by the following equations:
    \begin{subequations}
       \label{eq:mpxgat_mat_h}
       \begin{align}
           e_{i,j}^{H_k} & = \text{LeakyReLU}\left(\left[\mW^{H_k} \cdot \vh_i^{H_k} \cdot (\vv^{H_k})^T|| \mW^{H_k} \cdot \vh_j^{H_k} \cdot (\vv^{H_k})^T\right]\right) \label{subeq:attention_coefficient_1_h}\\
           \alpha_{i,j}^{H_k} & = \text{softmax($e_{i,j}^{H_k}$)} = \frac{\exp(e_{i,j}^{H_k})}{\sum_{z \in \mathcal{N}_i^{H_k}} \exp(e_{i,z}^{H_k})} \label{subeq:attention_coefficient_2_h}\\
           \vh_i^{'H_k} & = \sigma \left( \sum_{j \in \mathcal{N}_i^{H_k}} \alpha_{i,j}^{H_k} \cdot \mW^{H_k} \cdot \vh_j^{H_k} \right) \label{subeq:embedding_framework_h}
       \end{align}
    \end{subequations}
    Equations \ref{eq:mpxgat_mat_h} govern the convolutional layer as described in the GAT model \citep{velivckovic2017graph}. 
    In particular, in Equation~\ref{subeq:embedding_framework_h}, $\vh'^{H_k}_i$ represents the horizontal embedding of the node \emph{i} that belongs to the layer $k$, while Equation~\ref{subeq:attention_coefficient_2_h} describes the self attention mechanism in layer $k$.

    Leveraging the horizontal embeddings, MPXGAT-V generates the vertical embeddings by making use of a custom mechanism inspired by the one applied in the GAT model, which we call GAT-V. 
    The equations to compute the vertical embedding of node \emph{i} are the following:
    \begin{subequations}
    \label{eq:mpxgat_mat_v}
       \begin{align}
        e_{i,j}^{V^{k,q}} &= \text{LeakyReLU}\left(\left[\mW^{V} \cdot \vh_i^{V} \cdot (\vv^{V})^T || \mW^{V} \cdot \vh_j^{V} \cdot (\vv^{V})^T \right]\right)  \label{subeq:attention_coefficient_1_v}\\
          \alpha_{i,j}^{V^{k,q}}    & =\text{softmax($e_{i,j}^{V^{k,q}}$)} =\frac{\exp(e_{i,j}^{V^{k,q}})}{\sum_{z \in \mathcal{N}_i^{V}} \exp(e_{i,z}^{V})} \label{subeq:attention_coefficient_2_v}                                  \\
          \vh_i^{'V} & =\sigma\left(\sum_{j\in\mathcal{N}_i^{V}}g\left(\left[\alpha_{i,j}^{V}\cdot \mW^{V}\cdot\vh_j^{V}\right], f(\vh_i^{H_k},\vh_j^{H_q})\right)\right) \label{subeq:embedding_framework_v}
       \end{align}
    \end{subequations}
    Equations~\ref{eq:mpxgat_mat_v} contain two novel elements. 
    The first is the function $f$, which transforms the final horizontal embeddings to the same dimensional space of the vertical embeddings.
    Such a function can be a simple linear transformation of the node horizontal embeddings, or it can be more complex, depending on the specific task considered.
    The second feature we introduce is the function $g$, which combines the data obtained from the current vertical embedding with the results of the the function $f$.    
    
    \subsection{MPXGAT Implementation for Link Prediction}
    We now present an implementation of \mpxgat{} that is suitable for link prediction. 
    We customize the attention mechanisms of both \mpxgatHmodel{} and \mpxgatVmodel{}, which are described in Equations~\ref{subeq:attention_coefficient_1_h}, \ref{subeq:attention_coefficient_2_h} and \ref{subeq:attention_coefficient_1_v}, \ref{subeq:attention_coefficient_2_v}, respectively. 
    In particular:
    \begin{itemize}
    \item in Equations~\ref{subeq:attention_coefficient_1_h} and~\ref{subeq:attention_coefficient_1_v}, instead of a single weight matrix for both $i$ and $j$ nodes, i.e., $\mW^H$, two different weight matrices, $\left(\mW^H_i,\mW^H_j\right)$, are used. 
    By doing that, we improve the representation capabilities of the model, at the expense of a greater number of parameters. 
    The same choice is done for the weight vector, where instead of a single vector $\vv^H,$ we use two, $\left(\vv^H_i,\vv^H_j\right)$. 
    Moreover, to improve the generalization capabilities of the model, the concatenation operator $||$ is implemented as a sum $+$, and both parts of this operation are augmented with the introduction of a bias vector;
    \item in Equations~\ref{subeq:attention_coefficient_2_h} and~\ref{subeq:attention_coefficient_2_v} an additional dropout mechanism is applied to the attention coefficients. In this way, we sub-sample the paths present in the graph, granting the model a better generalization capability. 
    \end{itemize} 
    
    A further addition is the use of \emph{multiple attention heads}, consisting in calculating different embeddings of the same nodes applying multiple times the convolutional layer.
    This generates multiple node representations that highlight different kinds of patterns about the nodes, as each attention head can be seen as a distinct information channel representing a certain aspect of nodes.
    The implemented sub-models concatenates the embeddings for each attention head in the hidden convolutional layers of the model, and uses a mean operation for the final one. 
    
    Finally, for both sub-models the activation function $\sigma$ in Equations~\ref{subeq:embedding_framework_h} and \ref{subeq:embedding_framework_v} we consider the \emph{LeakyReLU}.

    The general \mpxgat{} framework rely on the $f$ and $g$ functions for the embedding, which can be specified according to the particular task we want to solve.
    Here, the two functions $f$ and $g$ introduced in Equation~\ref{subeq:embedding_framework_v} have the following characteristics:
    \begin{itemize}
        \item $f$ performs a linear transformation of the horizontal embedding of the source node, $\vh_i^H$, to the same dimensional space of its vertical embedding, $\vh_i^V$, here described with $\vx_i^H$. 
        This is done through an additional weight matrix $\mZ^H$. 
        The result is then multiplied by a vector of parameters $\vv_i^H$, whose role is to enhance the patterns from the horizontal embeddings. 
        Finally, the LeakyReLU function is applied. 
        \item The function $g$ performs a weighted sum of the output of the function $f$ (see Equation~\ref{subeq:mpxgat_v_f_3}) and the data computed with the convolutional layer for the vertical network (see Equation~\ref{subeq:mpxgat_v_g_1}). 
        A scalar parameter $\beta$, automatically inferred during the learning process, is introduced to tune the information coming from the horizontal and vertical embeddings, respectively.
        In particular when $\beta = 0$ the horizontal embedding is not considered, while for $\beta = 0.5$ both the components are considered with the same weight. 
        
    Formally, the equations describing the implementation of \mpxgat{} for link predictions are:
    \end{itemize}
    \begin{subequations}
        \label{eq:mpxgat_v_f_g}
        \begin{align}
            \vx_i^H & = \vh_i^H \cdot \mZ^{H} + \vb_{h_i} \label{subeq:mpxgat_v_f_1}\\
            \alpha_i^H & = \text{LeakyReLU}\left(\ \vx_i^H \cdot \vv_i^H\right) \label{subeq:mpxgat_v_f_2}\\
            \vm^H_i & = f(\vx_i^H) = \alpha_i^H * \vx_i^H \label{subeq:mpxgat_v_f_3}\\
            \vm^V_{i,j} & = \alpha_{i,j}^V \cdot \vx_j^V \label{subeq:mpxgat_v_g_1}\\
            g\left(\vm^V, \vm^H\right) & = \left(\vm^V_{i,j}*\left( 1 - \text{ReLU}\left(\beta\right) \right)\right) + \left(\left(\vm^H_i\right) * \text{ReLU}\left(\beta\right)\right) \label{subeq:mpxgat_v_g_2}
        \end{align}
    \end{subequations}

A few aspects of our model are worth remarking.
First, we assume neither to have the same number of nodes in each horizontal layer nor to have all possible inter-layer links, making our algorithm a valuable tool for reconstructing inter-layer connectivity.
Second, \mpxgat{} is quite flexible, as it supports the use of node features and edge weights.
In particular, different features can be used by the two sub-models, since the horizontal and the vertical networks are defined and processed as separated entities.   
As we will show in the next section, these characteristics guarantee our model good performances even in datasets where the edge density in the horizontal or vertical networks is high, which represents a strong limit of algorithms based on GraphSAGE \citep{gallo2023multiplexsage}.

    \section{Results}
    In this section, we present the results obtained by \mpxgat{} in predicting both intra-layer and inter-layer links.
    We first present the datasets studied, as well as the experimental setup used for the analysis, and the competing methods considered as benchmarks for our algorithm.
    Then, we test the performances of \mpxgat{} in the link prediction task.
    We round up the analysis by assessing the impact of the horizontal and vertical sub-models on the embedding procedure.
    
    \subsection{Dataset}
    To assess the performance of \mpxgat{} in predicting both intra-layer and inter-layer connections, we employ the same datasets that have been originally used to test \multiplexsage{} \citep{gallo2023multiplexsage}. 
    These include data relative to three types of real-world multiplex networks, namely a collaboration, a biological, and an online social networks.
    Specifically, we considered the following datasets.
    
    \textbf{arXiv} \citep{de2015arxiv}. 
    The arXiv multiplex network represents collaborations in various research topics published on the pre-print archive. 
    Each layer of the network corresponds to a different research category or theme. 
    The network was obtained by selecting papers published before May 2014 that contain the keyword \emph{networks} in their titles or abstracts.
    
    \textbf{Drosophila} \citep{stark2006biogrid}. 
    This multiplex network represents the interactions between proteins and genes in the common fruit fly, i.e., Drosophila melanogaster, with each layer corresponding to interactions of various types. 
    The dataset was collected from the Biological General Repository for Interaction Datasets (BioGRID), with data updated until January 2014.
    
    \textbf{ff-tt-yt} \citep{Dickison2016MultilayerSN}. This multiplex network is derived from Friendfeed (ff), a social media aggregation platform where users can link their accounts from various online social networks (OSNs). 
    The network comprises users who have registered a single Twitter (tw) account and a sole YouTube (yt) account on Friendfeed. 
    Additionally, the Twitter and YouTube accounts are linked to a single Friendfeed account.
    
    For each empirical dataset, we consider exclusively the largest connected component of the multiplex networks and we treat all networks as undirected and unweighted. 
    Details about the largest connected component for each of these datasets are provided in Table \ref{tab:dataset-info}.
    
    We remark that, for all datasets considered in our analysis, nodes are not provided with any external features.
    Hence, we associate to each node, $n$, a one-hot encoding vector defined by the Kronecker function, $\delta_{i,n}$.
    
    \begin{table}[h!]
        \caption{Dataset Information including the number of nodes, edges, and average degree for the largest connected component of each dataset (arXiv, Drosophila, and ff-tw-yt)}
        \label{tab:dataset-info}
        \begin{center}
            \begin{tabular}{cccc}
                \textbf{Dataset} & \textbf{Nodes} & \textbf{Edges} & \textbf{Avg. Degree} \\
                \hline\hline
                arXiv & 19,310 & 20,738 & 1.07 \\
                Drosophila & 11,867 & 5,171 & 0.44 \\
                ff-tw-yt & 11,827 & 6,028 & 0.51 \\
                \hline
            \end{tabular}
        \end{center}
    \end{table}
    
    \subsection{Experimental setup}
    To assess the performance of \mpxgat{}, we consider the experimental setup delineated in \cite{gallo2023multiplexsage}. 
    We partition the data following a multiple step procedure.
    First, we randomly select 20\% of the network nodes, labeling them as \emph{marked nodes}. 
    We then define test and training sets. 
    Both sets encompass positive and negative examples, the former corresponding to actual links within the network, while the latter consisting in pairs of unconnected nodes.
    In the test set, positive examples comprise a subset of 20\% of the intra-layer links and all inter-layer links among the marked nodes. 
    Conversely, positive examples in the training set include all remaining intra-layer and inter-layer links in the multiplex network.
    As negative examples within the test set, we include 20\% of all possible negative intra-layer links among the marked nodes, as well as all possible inter-layer links between them. 
    The negative examples in the training set correspond to the remaining pairs of unconnected nodes.
    
    To test the performance of our algorithm, instead of conducting a single experiment for each dataset, we repeat the embedding procedure multiple times. 
    For each repetition, we randomly select a subset of marked nodes, defining the training and test sets accordingly.
    To obtain the best parameter settings, we apply a grid search method.
    
    \subsection{Competing Methods}
    We conduct a comparative analysis of \mpxgat{} against three competing methodologies, specifically \graphsage{}, \gatne{} and \multiplexsage{}.
    
    \graphsagebf{} \citep{hamilton2017inductive}. 
    \graphsage{} is an inductive node embedding algorithm that leverages node features to learn an embedding function capable of generalizing to unseen nodes. 
    It was originally designed for single-layer network embeddings, so in our experiments, we apply it without distinguishing between intra-layer and inter-layer links.
    
    \gatnebf{} \citep{cen2019representation}. 
    \gatne{} is an embedding algorithm for attributed heterogeneous networks, encompassing multigraphs with diverse node and edge types. 
    To adapt \gatne{} to our specific task, we introduced two categories of edges, denoted as intra-layer and inter-layer links. 
    This adjustment allows us to create a multigraph that the algorithm can learn to embed effectively.
    
    \multiplexsagebf{} \citep{gallo2023multiplexsage}. \multiplexsage{} represents an extension of the GraphSAGE algorithm, specifically tailored for embedding multiplex networks. 
    Its key feature is the distinction made between inter-layer and intra-layer links that allows for the prediction of both intra-layer and inter-layer connectivity patterns.
    
    Other relevant methods such as \citep{gong2020heuristic}, \citep{ioannidis2020tensor}, \citep{liu2017principled}, \citep{qu2017multinet}, \citep{zhang2018scalable}, \citep{shi2018mvn2vec} and \citep{huang2020mr} can not be used as competing methods, as they assume that each layer has the same number of nodes and that all inter-layer connections are known. 
    Therefore, as we are interested in predicting both intra-layer and inter-layer links, these methods are not suitable benchmarks for evaluating the performance of \mpxgat{}.
    
    \subsection{Embedding Multiplex Networks}
    Our first analysis concerns the prediction of both intra-layer and inter-layer links when embedding a multiplex network. 
    For each embedding, we evaluate the Area Under the Receiveing Operating Characteristic (ROC) Curve (AUC), and take an average over the different repetitions as a performance metric.
    We consider the standard deviation as an indicator of statistical error.
    Table \ref{tab:roc_auc_results_multiplexsage_cut} provides an overview of the performances obtained with \mpxgat{}, \graphsage{}, \gatne{}, and \multiplexsage{}.
    We note that \gatne{} outperforms \graphsage{} and \multiplexsage{} in intra-layer link prediction, having a higher average AUC for all three datasets.
    However, \mpxgat{} has comparable performances with \gatne{} on the ff-ww-tt and the Drosophila dataset, while the latter performs better on the arXiv dataset.
    Yet, with regards to inter-layer link prediction, \mpxgat{} clearly performs better than all other algorithms, including \multiplexsage{}, which is explicitly designed for that task.
    Overall, if we consider the general performance of the methods without distinguishing intra- and inter-layer connections, our algorithm emerges as the best solution, as reported in Table \ref{tab:cumulative_roc_auc_results_multiplexsage_cut}.
    
    \begin{table}[h!]
        \caption{Performance comparison on intra-layer and inter-layer link prediction across the three distinct datasets: ff-tw-yt, Drosophila, and arXiv. The assessment is based on the AUC metric, using the standard deviation as error metric.  
        The best performing tool is highlighted in boldface}
        \label{tab:roc_auc_results_multiplexsage_cut}
        \small 
        \centering
        \renewcommand{\arraystretch}{1.2} 
        \resizebox{\linewidth}{!}{
            \begin{tabular}{c|cc|cc|cc}
                \textbf{Algorithm} & \multicolumn{2}{c|}{\textbf{ff-tw-yt}} & \multicolumn{2}{c|}{\textbf{Drosophila}} & \multicolumn{2}{c}{\textbf{arXiv}} \\
                & \emph{intra} & \emph{inter} & \emph{intra} & \emph{inter} & \emph{intra} & \emph{inter} \\
                \hline\hline
                \textbf{\graphsage{}} &0.47 (± 0.02) &0.56 (± 0.02) &0.54 (± 0.02) &0.63 (± 0.02) &0.72 (± 0.02) &0.70 (± 0.01) \\
                
                \textbf{\gatne{}} &\textbf{0.83 (± 0.01)} &0.47 (± 0.01) &\textbf{0.78 (± 0.01)} &0.55 (± 0.01) &\textbf{0.91 (± 0.01)} &0.63 (± 0.01) \\
                
                \textbf{\multiplexsage{}} &0.48 (± 0.02) &0.62 (± 0.02) &0.51 (± 0.01) &0.77 (± 0.02) &0.71 (± 0.02) &0.83 (± 0.01) \\
                
                \textbf{\mpxgat{}} &0.76 (± 0.06) &\textbf{0.83 (± 0.01)} &0.76 (± 0.05) &\textbf{0.86 (± 0.02)} &0.80 (± 0.02) &\textbf{0.84 (± 0.01)} \\
            \end{tabular}%
        }
    \end{table}

    \begin{table}[h!]
        \caption{Overall AUC performaces calculated as a weighted sums of the results shown in Table \ref{tab:roc_auc_results_multiplexsage_cut}, based on the number of edges used to evaluate the models.}
        \label{tab:cumulative_roc_auc_results_multiplexsage_cut}
        \small 
        \centering
        \renewcommand{\arraystretch}{1.2} 
            \begin{tabular}{c|c|c|c}
                \textbf{Algorithm} & \textbf{ff-tw-yt} & \textbf{Drosophila} & \textbf{arXiv} \\
                \hline\hline
                \textbf{\graphsage{}} &0.49 (± 0.02) & 0.57 (± 0.01) & 0.70 (± 0.01) \\
                
                \textbf{\gatne{}} & 0.72 (± 0.01) & 0.69 (± 0.02) & 0.72 (± 0.01) \\
                
                \textbf{\multiplexsage{}} & 0.52 (± 0.02) & 0.61 (± 0.01) & 0.79 (± 0.01) \\
                
                \textbf{\mpxgat{}} & \textbf{0.78 (± 0.03)}  & \textbf{0.80 (± 0.03)} & \textbf{0.82 (± 0.04)} \\
            \end{tabular}
    \end{table}
    
    As described in Section \ref{sec:methods}, \mpxgat{} employs two distinct embeddings, dealing with the horizontal, i.e., intra-layer, and vertical, i.e., inter-layer, embeddings, respectively. 
    In contrast, the other models rely on a single embedding that serves both tasks. 
    As we will further investigate in the next section, it is this architectural difference that allows \mpxgat{} to predict with a certain reliability both intra-layer and inter-layer links.

    \subsection{Measure the impact of Horizontal Embeddings}
    We now establish the impact of the horizontal embeddings on the performance of \mpxgat{}.
    To do so, we conduct two experiments.
    In the first, we perform the embedding with \mpxgat{}, but instead of using MPXGAT-V, which leverages the horizontal embeddings generated by MPXGAT-H, for embedding the vertical network, we consider a standard GAT \citep{velivckovic2017graph}, thus ignoring the contribution of MPXGAT-H.
    Table \ref{tab:ablation_tests_no_horizontal_embeddings} reports the average AUC for the inter-layer link prediction in both configurations. 
    We observe that neglecting the horizontal embeddings leads to worse performances across all datasets, suggesting that including the MPXGAT-V submodel increases the algorithm adaptability and predictive power.
    
    To validate the significance of these findings we performed a Welch's T-test.
    The p-values are $6.00 \times 10^{-7}$, $9.10 \times 10^{-10}$, and $1.10 \times 10^{-7}$ for the Drosophila, arXiv, and ff-tw-yt datasets, respectively, confirming the statistical significance of our finding.
    
    \begin{table}[h!]
        \caption{Results of the ongoing experiment aimed at assessing the impact of excluding horizontal embeddings on inter-layer link prediction. The performance is evaluated in terms of AUC across the three datasets. The second model, which omits horizontal embeddings, exhibits lower performance in inter-layer link prediction across all datasets.
        }
        \label{tab:ablation_tests_no_horizontal_embeddings}
        \centering
        \scalebox{0.9}{
        \begin{tabular}{ccccc}
            \textbf{Algorithm} & \textbf{ff-tw-yt} & \textbf{Drosophila} & \textbf{arXiv}\\
            \hline
            \mpxgatVmodel \emph{(layer \gatv{})} & \textbf{0.83 ± (0.01)} & \textbf{0.86 ± (0.01)} & \textbf{0.84 ± (0.01)} \\
            \gat \emph{(layer \gat)} & 0.72 ± (0.02) & 0.78 ± (0.02) & 0.78 ± (0.01)\\
            \hline
        \end{tabular}
        }
    \end{table}
    
    In the second experiment, we keep the original model architecture but instead of providing MPXGAT-V with the horizontal embeddings generated by MPXGAT-H, we replace them with random embeddings, i.e., vectors whose components are random values.
    
    Table \ref{tab:ablation_tests_non_significative_embeddings} shows the results of the comparison with regards to the inter-layer link prediction. 
    We note that for both the ff-tw-yt and the arXiv dataset, using the horizontal embeddings increases the performance of the prediction task, while we observe similar values of the average AUC for the Drosophila dataset. 
    The statistical significance of this result is confirmed by the Welch's T-test, for which we obtain p-values equal to $6.10 \cdot 10^{-6}$, $0.75$, $3.60 \cdot 10^{-4}$ for the arXiv, Drosophila, and ff-ww-tt datasets, respectively. 
    
    We conjecture that this outcome is due to the structure of the multiplex network, both in terms of the size of the different layers, i.e., the number of nodes laying on them, and of the connectivity patterns, both within the same layer (e.g., randomness, clustering and community organization, and across different ones, i.e., overlapping).
    
    \begin{table}[h!]
        \caption{Results of the experiment investigating the impact of replacing meaningful horizontal embeddings with random embeddings. The performance is evaluated in terms of AUC. The model with random embeddings performs worse in two out of the three datasets compared to the model with actual embeddings, indicating a significant decrease in predictive accuracy.}
        \label{tab:ablation_tests_non_significative_embeddings}
        \centering
        \scalebox{0.85}{
        \begin{tabular}{lcccc}
            \textbf{Algorithm} & \textbf{ff-tw-yt} & \textbf{Drosophila} & \textbf{arXiv}\\
            \hline
            \mpxgatVmodel \emph{(actual embedding)} & \textbf{0.83 ± (0.01)} & 0.86 ± (0.01) & \textbf{0.84 ± (0.01)} \\
            \mpxgatVmodel \emph{(random embedding)} & 0.80 ± (0.02) & 0.86 ± (0.01) & 0.81 ± (0.01)\\
            \hline
        \end{tabular}
        }
    \end{table}
    
    \section{Conclusions}
    In this paper we have introduced \mpxgat{}, an attention based deep learning model for the embedding of multiplex graphs. 
    Through a comprehensive experimental analysis we showed that MPXGAT out-performs state-of-the-art competing algorithms.
    Future work will be aimed at better understanding how the community structure withing each layer influences the performances of the algorithm in terms of the reliability of predicted inter-layer relations.
    An implementation of \mpxgat{} in Python is available at https://github.com/MarcoB46/MPXGAT.

\bibliographystyle{elsarticle-num} 
\bibliography{biblio}
\end{document}

%% file: math_commands.tex
\usepackage{amsmath,amsfonts,bm}

\def\eqref#1{equation~\ref{#1}}

\def\1{\bm{1}}

\def\vb{{\bm{b}}}

\def\vh{{\bm{h}}}

\def\vm{{\bm{m}}}

\def\vv{{\bm{v}}}

\def\vx{{\bm{x}}}

\def\mW{{\bm{W}}}

\def\mZ{{\bm{Z}}}

\DeclareMathAlphabet{\mathsfit}{\encodingdefault}{\sfdefault}{m}{sl}
\SetMathAlphabet{\mathsfit}{bold}{\encodingdefault}{\sfdefault}{bx}{n}

\def\gG{{\mathcal{G}}}

\def\sV{{\mathbb{V}}}

\newcommand{\R}{\mathbb{R}}

